\pdfoutput=1
\documentclass[11pt]{article}

\usepackage{ACL2023}

\usepackage{times}
\usepackage{latexsym}
\usepackage{graphicx}
\usepackage{float}
\usepackage{makecell}

\usepackage[T1]{fontenc}

\usepackage[utf8]{inputenc}

\usepackage{microtype}

\usepackage{inconsolata}

\title{Multiclass Sentiment Analysis for Identifying Political Viewpoints}

\author{ Girma Yohannis Bade,  Olga Kolesnikova, Jose Luis Oropeza, Grigori Sidorov \\ Centro de Investigaciones en Computación (CIC), Instituto Politécnico Nacional (IPN), \\ Miguel Othon de Mendizabal, Ciudad de México, 07320, México }

\begin{document}
\maketitle
\begin{abstract}
The rapid growth of social media has created vast amounts of political discourse, which provides valuable opportunities to analyze public opinions and identify different political perspectives. Sentiment Analysis (SA) is a core task in Natural Language Processing (NLP) that allows the computational study of attitudes and opinions in textual data, and has become increasingly important for understanding political discourse. In this work, we investigate multiclass sentiment analysis of political viewpoints on social media, that is to automatically discriminate multiple sentiment classes over political issues and figures. To solve this task we design and evaluate two machine-learning approaches based on XGBoost and BERT. We train and evaluate the models on a labeled dataset of political social media posts using standard classification metrics. The experimental results show that the XGBoost model reaches an F1-score of 0.2835 and the BERT-based model reaches an F1-score of 0.2806 on the test set. These results demonstrate the challenge of classifying complex and contextualized political discourse sentiment and provide a baseline for future research in multiclass political sentiment analysis.

\end{abstract}
\begin{k}
\end{k}

\section{Introduction}
Social media has gained increasing popularity and is a valuable source of diverse, expressive, and real-time political discourse, especially X (former Twitter). These platforms provide an opportunity for users to freely express their thoughts, feelings, opinions and comments on a variety of topics. Sentiment analysis (SA) plays an important role in areas such as political affairs, social awareness, international conflicts, movie reviews, education systems, and feedback on products \cite{bade2026evaluating}. It enables the extraction of valuable insights from tweets to understand public sentiment and opinions, and has consequently been increasingly adopted by companies, government agencies, and other organizations to correct their stands \citep{ramanathan2021sentiment, kolesnikovadetecting}.

SA is a fundamental task in natural language processing (NLP) that aims to identify and classify opinions and attitudes expressed in textual data into predefined categories. In the political domain, sentiment analysis is particularly valuable for understanding public opinion, capturing diverse political perspectives, identifying societal concerns, and supporting evidence-based policymaking. Beyond politics, the ability to automatically identify sentiments toward products, organizations, services, and other entities has become increasingly important across a wide range of applications \citep{mullen2006preliminary, yigezuodio}.

For instance, \citet{mullen2006preliminary, yigezu2024habesha} demonstrated the potential of NLP-based sentiment analysis for examining political trends, complementing traditional opinion polls, identifying political bias in news and other ostensibly objective texts, and characterizing the views associated with particular texts and individuals. Such analyses can provide valuable insights for targeted communication and outreach activities, including political campaigns, public engagement, donation appeals, and petition initiatives.

Building on these applications, this paper presents a multiclass sentiment analysis framework for identifying and classifying political viewpoints expressed in social media discourse for Tamil language.
Most speakers of Dravidian languages (Tamil, Kannada, Malayalam, Telugu, Tulu and others) are found in South India and Sri Lanka represent a rich linguistic and cultural heritage of the region. The workshop is organized by DravidianLangTech to promote research in tackling the challenges arising from the lack of progress in language technology for these languages.
Every year, it provides a gold standard dataset for the selected tasks and organizes a workshop \cite{duraphe2022dlrg,coelho2023mucs,bade2024social, mersha2024semantic}. 

 We leveraged dataset provided for DravidianLangTech@NAACL 2025 \cite{bade2025girma} to classifying the political opinions expressed in Tamil tweets into seven distinct classes, such as: 1) Substantiated, 2) Sarcastic, 3) Opinionated, 4) Positive, 5) Negative, 6) Neutral, and 7) None of the above.
the main contribution of this paper is summarized as follows:
\begin{itemize}
\item We review the existing literature on sentiment analysis in the political domain and examine how sentiment analysis can be applied to monitor and analyze political discourse.

\item We investigate and select appropriate feature extraction methods for the given datasets and prepare the resulting representations for model training.

\item We select state-of-the-art language models, fine-tune them on the target datasets, and evaluate their performance on held-out test data using standard evaluation metrics.
\end{itemize}

 The paper is structured as follows: Section 2 discusses the related works. Section 3 mentions the methodology including data sets, approaches and experimentation. Finally Section 4 discusses the results obtained from the experiments.

\section{Related works}
SA is an interesting area of research in NLP \cite{bade2024social,bade2024hope,yigezu2024ethio,bade2024evaluating}. It is about a classification task that categorizes or predicts the linguistic input features based on the patterns trained through AI algorithms during the training phase. The concept can be adopted for numerous languages while employing AI approaches such as transformer-based, deep learning, and machine learning \cite{yigezu2023transformer}.

According to \citet{chan2023state}, many sentiment analysis problems, including as emotion detection, cross-domain sentiment classification, multimodal sentiment analysis, aspect-based sentiment analysis, and multilingual sentiment analysis, are useful for knowledge adaptation strategies. 
Basically, sentiment analysis is a broad umbrella for many NLP downstream tasks like hope speech, abusive and hate detection, stress identification, emotional analysis, and so on \cite{yigezu2023habesha,yigezu2023multilingual}.

For instance,\citet{ghosh2023multitasking, mersha2025evaluating}, attempted the first sentiment and emotional state recognition challenge, which triggered other researchers to contribute in under-resourced languages. A transformer-based multitask framework has been utilized by them to identify emotions and detect sentiment in code-mixed datasets. 
SA is not limited to text datasets; it can also operate on custom datasets that include political and film reviews. Code-mixed speech-sentiment classification has also been attempted in \cite{keshav2023multimodal}, utilizing the 3-shot, few-shot learning (FSL) framework and a fully connected neural network (FCNN) model. Although sentiment analysis (SA) is widely researched for languages with abundant resources\cite{bade4development, yigezu2023bilingual}, developing trustworthy systems for low-resource languages is still difficult because there is a lack of training data for this kind of work\cite{bade2018object}.
\section{Methodology}
This section provides comprehensive details about the dataset and the experimental settings adopted in this study. It also describes the overall system architecture, data encoding and preprocessing procedures, as well as the data format and configuration used for the selected models.

\subsection{Datasets}
For research in the NLP domain,  a well-articulated collection of datasets are driving fuel to produce insightful language models. For this study, we utilize the dataset collected as part of the DravidianLangTech 2025 shared task \citep{roy2025lexilogic}. The dataset is organized into three subsets: training, development, and test sets. The training and development sets are accompanied by corresponding sentiment labels, whereas the test set is unlabeled. A summary of these dataset splits is presented in Table~\ref{tbl1}.

\begin{table}[H]
\centering % Center the table
\caption{Dataset Statistics}
\label{tbl1} % The label should be placed after \caption
\begin{tabular}{|c|l|c|} % 'l' for left-aligned column instead of 'c' for datasets
\hline
\textbf{No} & \textbf{Datasets}              & \textbf{Sample Sizes} \\ \hline
1           & Train Dataset                  & 4,352                \\ \hline
2           & Development Dataset            & 544                 \\ \hline
3           & Test Dataset (Unlabeled)       & 544                   \\ \hline
\end{tabular}
\end{table}

As we can see from the Table \ref{tbl1}, it consisted three datasets. The training dataset is main dataset we use it for training our chosen algorithm. In case when we choose a supervised machine learning algorithm, it learns the pattern from this training data. The second set is development dataset, which is used to tune the behavior of our model during experimentation. It's mostly used to validate the model performance before applying the test data. In works like shared task where the model performance is tested by third party (organizer), validating the model with the development dataset gives the confidence before sending the final test predictions to the organizer. The test dataset is one that determines the final performance of the model. This data is  separate and never been seen during training. The Table \ref{tbl2} shows the class label distribution of training and development datasets.

\begin{table}[H]
\centering
\caption{Class label distribution  statistics}
\label{tbl2}
\begin{tabular}{|c|l|c|}
\hline
\textbf{Labels} & \makecell{\textbf{\# Count}\\\textbf{in Train}} & \makecell{\textbf{\# Count}\\\textbf{in Dev}} \\ \hline
Opinionated      & 1,361                                        & 153                                       \\ \hline
Sarcastic        & 790                                          & 115                                         \\ \hline
Neutral          & 637                                          & 84                                         \\ \hline
Positive         & 575                                          & 69                                       \\ \hline
Substantiated    & 412                                          & 52                                          \\ \hline
Negative         & 406                                          & 51                                          \\ \hline
None of the above & 171                                         & 20                                        \\ \hline
\end{tabular}
\end{table}

\subsection{Preprocessing}
The training, development, and test datasets were subjected to a standardized preprocessing procedure to improve the consistency and quality of the input data. The primary objectives of this preprocessing step were to remove punctuation marks, emojis, and user mentions, which may introduce noise or contribute limited semantic information in wrong way \cite{bade2025amado}. In particular, user mentions were removed to minimize the influence of author-specific or account-specific information on model predictions. The built-in Python re (regular expression) module was employed to identify and remove usernames and punctuation marks from the text. Emojis were also removed to ensure a consistent textual representation across the datasets. The same preprocessing procedure was applied uniformly to all dataset splits to maintain consistency between training, development, and test data and to prevent discrepancies arising from different preprocessing strategies.
 
\subsection{Feature Extraction}
Since practically all AI algorithms operate on numerical data, it is essential to appropriately encode language inputs into their corresponding numerical representations \cite{bade2024lexicon,bade2025pragmatic}. The process of transforming textual input into a numerical form is commonly referred to as data encoding or feature extraction \cite{bade2024lexicon}. Although various feature extraction techniques are available, we employed TF-IDF and BertTokenizer for the Logistic Regression and BERT models, respectively. This approach ensures that the input text is represented in a format compatible with the requirements of each model. Furthermore, the same feature extraction strategy was consistently applied across the relevant datasets to maintain comparability between the experimental settings.

\subsection{Model Selection}
Once the NLP part is ready, the next step is to choose and apply AI algorithms \cite{yigezu2023bilingual}. Thus, we have started our experiment with one of the traditional machine learning algorithms known as XGBClassifier. XGBoost is a widely used traditional machine learning algorithm that has demonstrated strong performance across a variety of classification tasks.
 To make it effective, we employed IF-IDF to extract the features from the text data.

In our second experiment, we chose the bert-base-uncased transformer model. For this model, there is its own BertTokenizer to tokenize and convert text data into numeric form. Table \ref{tbl3} presents its hyperparameters. 

\begin{table}[H]
    \centering
    \caption{BERT Hyperparameters}
    \label{tbl3}
    \begin{tabular}{|l|l|}
        \hline
        \textbf{Hyperparameters} & \textbf{Values}        \\ \hline
        Learning Rate            & 1e-5                  \\ \hline
        Evaluation Strategy      & Epoch                 \\ \hline
        Epochs                   & 5                     \\ \hline
        Batch Size               & 32                    \\ \hline
         Activation@output level               & softmax                    \\ 
        \hline
    \end{tabular}
\end{table}
As we can see from the table \ref{tbl3}, the learning rate indicates the number of times the execution took place to improve the model performance. The epoch refers to a complete pass through the entire training data set by the learning algorithm. Thus, we set the epoch to be 5,.i.e the execution did pass 5 complete times. The batch size refers to dividing the total data size into 32 and bringing the divided batch one a time for the execution. This helps the execution to be fast.
\begin{figure}[H]
    \centering
    \includegraphics[width=7.5cm,height=12cm]{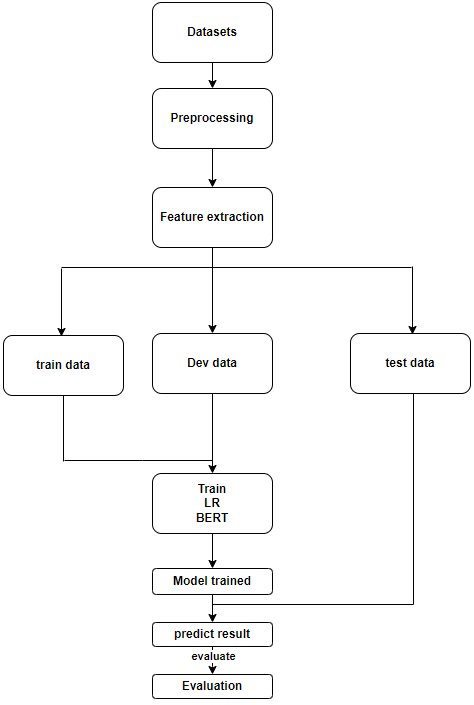}
    \caption{The work flow of proposed model}
    \label{fig1}
\end{figure}

\subsection{Results and Discussion}

The experimental results provide an empirical comparison of the machine learning and transformer-based approaches considered in this study. We first employed the XGBClassifier algorithm \cite{chang2022melanoma} to train the model on the provided training dataset and evaluated its performance on a separate test set. The XGBoost-based approach achieved a macro F1 score of 0.2835, indicating its performance in identifying the target classes under the given experimental setting.

Similarly, we conducted a second experiment using a BERT-based transformer model. The BERT model achieved a macro F1 score of 0.2806 on the same evaluation setting. The results of both approaches are presented in Tables \ref{tbl4} and \ref{tbl5}, respectively, providing a comparative view of their performance.
\begin{table}[H]
\centering
\caption{The result statistics of XGBboost on test data}
\label{tbl4}
\resizebox{0.5\textwidth}{!}{%
\begin{tabular}{|l|c|c|c|c|}
\hline
\textbf{Labels}             & \makecell{\textbf{Pr}} & \makecell{\textbf{Re}} & \makecell{\textbf{F1}} & \makecell{\textbf{Sup}} \\ \hline
Opinionated          & 0.1739      & 0.0870      & 0.1159      & 46
  \\ \hline
Sarcastic            & 0.1200      & 0.0857      & 0.1000      & 70  \\ \hline
Neutral              & 0.7500      & 0.7200      & 0.7347      & 25  \\ \hline
Positive             & 0.3607      & 0.6433      & 0.4622      & 171 \\ \hline
Substantiated        & 0.3256      & 0.1867      & 0.2373      & 75  \\ \hline
Negative             & 0.3095      & 0.2453      & 0.2737      & 106 \\ \hline
None of              & 0.1333      & 0.0392      & 0.0606      & 51 \\ \hline
\textbf{Accuracy}    & \multicolumn{4}{c|}{0.3309} \\ \hline
\textbf{Macro Avg}   & 0.3104      & 0.2867      & \textbf{0.2835}      & 544 \\ \hline
\textbf{Weighted Avg} & 0.2957      & 0.3309      & 0.2934      & 544 \\ \hline
\end{tabular}%
}
\end{table}

\begin{table}[H]
\centering
\caption{The BERT model on test data}
\label{tbl5}
\resizebox{0.5\textwidth}{!}{%
\begin{tabular}{|l|c|c|c|c|}
\hline
\textbf{Labels}             & \textbf{Pr} & \textbf{Re} & \textbf{F1} & \textbf{Sup} \\ \hline
Opinionated          & 0.2222      & 0.0870      & 0.1250      & 46  \\ \hline
Sarcastic            & 0.1250      & 0.0714      & 0.0909      & 70  \\ \hline
Neutral              & 0.7273      & 0.6400      & 0.6809      & 25  \\ \hline
Positive             & 0.3621      & 0.6374      & 0.4619      & 171 \\ \hline
Substantiated        & 0.2632      & 0.2667      & 0.2649      & 75  \\ \hline
Negative             & 0.3425      & 0.2358      & 0.2793      & 106 \\ \hline
None                 & 0.1429      & 0.0392      & 0.0615      & 51  \\ \hline
\textbf{Accuracy}             & \multicolumn{4}{c|}{0.3327}               \\ \hline
\textbf{Macro Avg}            & 0.3122      & 0.2825      & \textbf{0.2806 }     & 544 \\ \hline
\textbf{Weighted Avg}         & 0.2985      & 0.3327      & 0.2955      & 544 \\ \hline
\end{tabular}%
}
\end{table}
In Tables \ref{tbl4} and \ref{tbl5}, the column headings Pr, Re, F1, and Sup represent precision, recall, F1 score, and support, respectively. Support indicates the number of true instances in each labels. These metrics, along with accuracy, macro average, and weighted average, are used to evaluate performance. Among these, the macro average F1 score is often the most significant metric, as it relies on the harmonic mean of recall and precision, providing a balanced measure of performance. Therefore, our work is ranked based primarily on the F1 score values. 

\begin{table}[H]
\centering
\caption{The number of true instances and predicted instances' statistics in the labels.}
\label{tbl5}
\resizebox{0.5\textwidth}{!}{%
\begin{tabular}{|l|c|c|c|}
\hline
\textbf{Labels}             & \textbf{\#Actual} & \textbf{\#Predicted} & \textbf{Remark}\\ \hline
Opinionated          & 46        &   305    &   over    \\ \hline
Sarcastic            & 70      & 84      & over      \\ \hline
Neutral              & 25      &  50     &      over \\ \hline
Positive             & 171       &  43    &   under   \\ \hline
Substantiated        & 75       &   15    &   under   \\ \hline
Negative             & 106      &   23    &    under   \\ \hline
None                 & 51      & 24     & under       \\ \hline
\textbf{Total}                 & 544      & 544     & ---      \\ \hline
\end{tabular}%
}
\end{table}
Table \ref{tbl5} makes a figurative comparison between actual number of lables and predicted number of labels. For instance, the label 'Opinionated' had 46 instances in test set which are manually annotated but our model (XGBboost) made 305 prediction for it. Thus, the predicted values are greater than actual and hence it is marked as 'over' in the remark column. The reason for this might be the algorithm has learned more number of this class, see Table \ref{tbl2}. Similarly, the label 'Positive' had 171 instances but our model's prediction is 43. Therefore, the number of predicted instances are less than actual, and hence it marked as 'under'.

\begin{figure}[H]
    \centering
    \includegraphics[width=8.5cm,height=8cm]{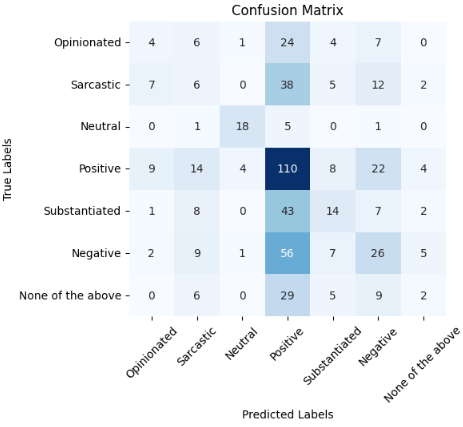}
    \caption{Confusion matrix that shows results of XGBboost algorithm}
    \label{fig2}
\end{figure}

\begin{figure}[H]
    \centering
    \includegraphics[width=7.5cm,height=8cm]{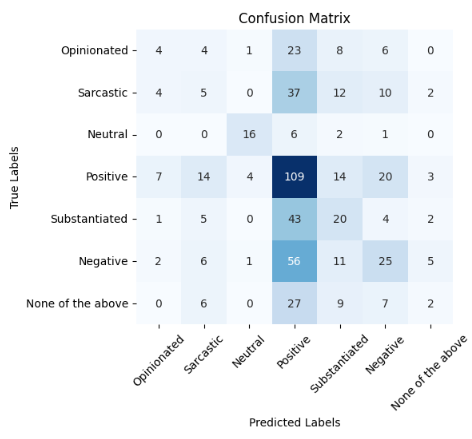}
    \caption{Confusion matrix that shows results of BERT model}
    \label{fig3}
\end{figure}

Figures \ref{fig2} and \ref{fig3} visualize the results in confusion matrix.

In Figure \ref{fig2}, the diagonal elements are corret predictions. For example the label 'Positive' has been correctly classified \textbf{110} times, holding the first position. Next, the label 'Negative' has been classified moderately 26 times. In third position, the label 'Neutral' is predicted 18 times correctly. On the other hand, the model misclassified the label 'Negative' as positive 56 times,holding first position. Next, 'Substantiated' is classified as 'positive' wrongly 43 times. In third osition, the label 'sarcastic' is misclassified as positive 38 times. Similarly, 'None of the above' and 'Opinionated' are misclassified as 29 times and 24 times respectively. Finally, the cell with 0 values indicate that the label in the XY coordinate have never been mixed up. For instance, the label 'Opinionated' has never been classified to 'None of the above' and vice versa. Likewise, 'Neutral' is not misclassified to 'Non of the above' and 'Non of the above'is not misclassified to 'Neutral'. All others that are not explicitly mentioned can be defined in similar fashion. 

In Figure \ref{fig3}, diagonal elements are correct predictions. For instance, 'Positive' is correctly classified the most (109 times), suggesting the model performs best in identifying Positive sentiment.
'Neutral' (16 times) and 'Negative' (25 times) have moderate correct classifications.
'None of the Above' label (2 times) is poorly classified, meaning the model struggles with identifying this category.
The most common misclassifications
'Opinionated' is misclassified as 'Positive' (23 times).
Sarcastic is frequently misclassified as 'Positive' (37 times) and 'Substantiated' (12 times).
'Substantiated' is also mostly misclassified as 'Positive' (43 times).
'Negative' has 56 cases misclassified as 'Positive'.
'None of the Above' is often confused with 'Positive' (27 times).

\section{Conclusion and Future Work}
In this task, we have developed social media classifying models and evaluated their performance using various metrics. The developed model is able to classify social media posts into  into seven multiclasses as expected. The two AI algorithms we employed here are XGBboost and Bert model. As our result show, XGBboost outperformed bert in this usecase.

As a direction for future research, similar studies should be conducted for a wider range of languages, particularly those that are underrepresented in existing research, since political opinions and public discourse are increasingly expressed through multilingual social media platforms. Extending the analysis to additional languages would provide a broader understanding of political opinion across different linguistic and cultural contexts. Furthermore, the performance of the proposed models could be improved by incorporating and comparing additional machine learning and deep learning algorithms for the languages considered in this study, as well as by increasing the size and diversity of the datasets. Such extensions could contribute to more robust, generalizable, and language-independent models for political opinion analysis.

\section*{Acknowledgments}
The work was done with partial support from the Mexican Government through the grant A1-S-47854 of CONACYT, Mexico, and grants 20241816, 20241819, and 20240951 of the Secretaría de Investigación y Posgrado of the Instituto Politécnico Nacional, Mexico. The authors thank the CONACYT for the computing resources brought to them through the Plataforma de Aprendizaje Profundo para Tecnologías del Lenguaje of the Laboratorio de Supercómputo of the INAOE, Mexico and acknowledge the support of Microsoft through the Microsoft Latin America PhD Award.

\section*{Limitation and Ethics Statement}
Since Tamil is a language with limited resources and the model was trained using a tiny dataset, the performance observed may not generalize well to all unseen data.  Despite the challenges of limited resources and competition, our model demonstrated strong performance in classifying multiclass political sentiments in Tamil social media posts. Furthermore, our work adheres to the ethical principles outlined for computational research and professional conduct\footnote{\url{https://www.aclweb.org/portal/content/acl-code-ethics}}.

\bibliography{custom}
\bibliographystyle{acl_natbib}
\end{document}